\documentclass[aps,prx,twocolumn,reprint,amssymb,floats,superscriptaddress]{revtex4-2}
\usepackage{graphicx} % Required for inserting images
\usepackage{color}
\usepackage{amsmath}
\usepackage{physics}
\usepackage{multirow}
\usepackage{hyperref}

\begin{document}

\title{Regularized second-order optimization of tensor-network Born machines}
\author{Matan Ben Dov}
\email{matan.ben-dov@biu.ac.il}
\affiliation{Department of Physics, Bar-Ilan University, 52900 Ramat Gan, Israel}
\affiliation{Zapata AI., Boston, United States}
%\affiliation{Center for Quantum Entanglement Science and Technology, Bar-Ilan University, 52900 Ramat Gan, Israel}
\author{Jing Chen}
\email{yzcj105@gmail.com}
\affiliation{Zapata AI., Boston, United States}
% \author{ Matan Ben-Dov, Jing Chen}
\date{\today}

\begin{abstract}
    % Tensor-network Born machine (TNBM) is a quantum-inspired generative model for learning data distribution \cite{han2018unsupervised}. Using tensor-network contraction and optimization techniques and the negative log-likelihood loss, the model learn an efficient representation of the distribution, capable of ...
    % Despite its promises, the optimization of TNBMs encounter different challenges when optimizing each tensor core individualy. Due to the logarithmic nature of the loss function, the single-tensor optimization problem cannot be solved analytically, requiring an iterative optimization which slows down the optimization process and is prone to converging to one of the many non-optimal local minima.
    % In this paper we present an improved second-order optimization technique for TNBM optimization, which provides a robust improvement in training rates and quality of the optimized model. Our method uses a modified Newton's method optimization on the manifold of normalized states, with the introduction of regularization terms to avoid local minima. 
    % We demonstrate our technique for a 1-dimensional matrix product state trained on discrete and continuous datasets, providing a ...
    
    Tensor-network Born machines (TNBMs) are quantum-inspired generative models for learning data distributions. Using tensor-network contraction and optimization techniques, the model learns an efficient representation of the target distribution, capable of capturing complex correlations with a compact parameterization. 
    Despite their promise, the optimization of TNBMs presents several challenges. A key bottleneck of TNBMs is the logarithmic nature of the loss function commonly used for this problem. The single-tensor logarithmic optimization problem cannot be solved analytically, necessitating an iterative approach that slows down convergence and increases the risk of getting trapped in one of many non-optimal local minima.
    
    In this paper, we present an improved second-order optimization technique for TNBM training, which significantly enhances convergence rates and the quality of the optimized model. Our method employs a modified Newton’s method on the manifold of normalized states, incorporating regularization of the loss landscape to mitigate local minima issues.     
    We demonstrate the effectiveness of our approach by training a one-dimensional matrix product state (MPS) on both discrete and continuous datasets, showcasing its advantages in terms of stability and efficiency, and demonstrating its potential as a robust and scalable approach for optimizing quantum-inspired generative models.
    
\end{abstract}
\maketitle

\section{Introduction}

\begin{figure}[t]
    \centering
    \includegraphics[width=0.95\linewidth]{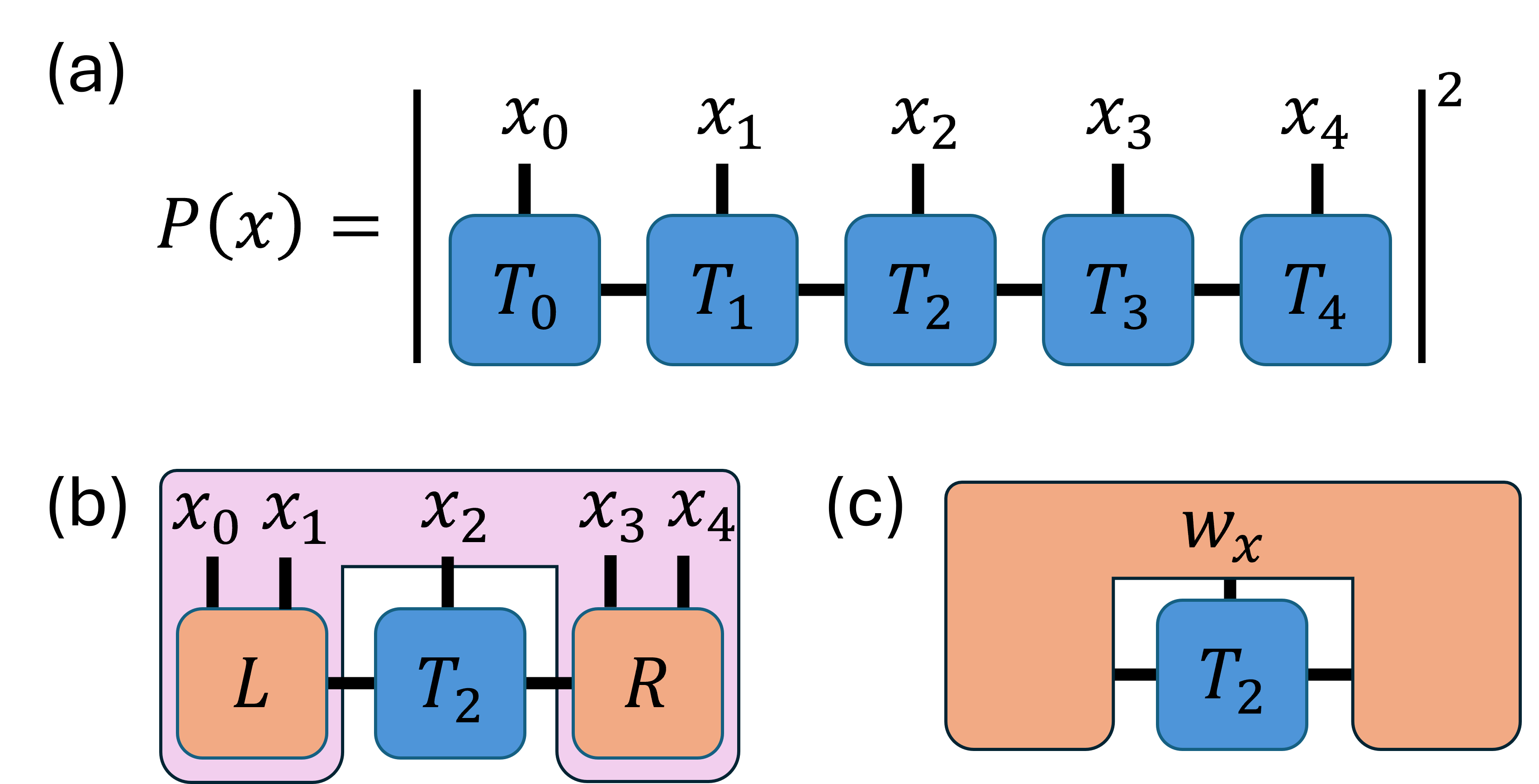}
\caption{Illustration of the matrix product state (MPS) and its application to TNBMs. Panel (a) shows the probability of a sample, represented as the square of the overlap between the binary bit-string of the sample and the MPS. The example illustrates a 5-site MPS. %The gradients w.r.t. a single tensor $T_2$ involves an essential step (panel (b)) of the contraction of environment tensor $w_x$ (panel c), which is carried out by contracting the left hand side tensor cores ($L$) and the right hand side tensor crores ($R$), together with one-hot encoding of the entire bitstring sample, which encapsulates the dependency of the global overlap on a single tensor core. 
Panel (b) depicts the contraction of the MPS into left-environment ($L$) and right-environment ($R$) tensors. These environment tensors, along with the bitstring sample, are further contracted to form the full environment tensor, as shown in panel (c). The full environment tensor plays a crucial role in the formulation of the gradient with respect to a single tensor ($T_2$).}
\label{fig:TN_illustration}
\end{figure}

Generative modeling lies at the heart of modern machine learning, enabling tasks such as image synthesis, text generation, and data augmentation. In recent years many models and training methods have been devised to efficiently learn complex distribution from a given set of training samples, such as Generative Adversarial Networks (GANs) \cite{goodfellow2014generative, gui2021review}, Variational Autoencoders (VAEs) \cite{kingma2019introduction} and Boltzmann machines \cite{ackley1985learning, salakhutdinov2009deep}, as well as transformer-based and diffusion-based models \cite{vaswani2017attention,lin2022survey, ho2020denoising, yang2023diffusion}.

Among the diverse array of generative models, tensor-network Born machines (TNBMs) have emerged as a promising quantum-inspired framework for generative modeling \cite{han2018unsupervised, cheng2019tree, sun2020generative, cheng2021supervised}. TNBMs capitalize on the mathematical framework of tensor networks to efficiently represent and generate complex, high-dimensional probability distributions. These models harness the power of tensor networks, such as matrix product states (MPS) and projected entangled pair states (PEPS), originally introduced in the context of quantum wave-functions, thereby enabling a compact yet expressive description of intricate probability distributions \cite{verstraete2008matrix, orus2014practical,montangero2018introduction, banuls2023tensor, garcia2024survey}.
The introduction of tensor networks as a tool for machine learning offers a tractable way to represent the inner correlation between parts of the bitstrings using tensor cores of controllable sizes, and can help the model generalize in a natural way \cite{strashko2022generalization,caro2022generalization}. 

% The optimization of tensor-network based model is commonly carried by a tensor-by-tensor approach sweeping over all tensor cores and optimizing a single tensor core at a time. For one dimensional tensor networks, known as matrix product state (MPS) or tensor trains (TT), optimization is often interleaved with gauge-fixing procedure, moving the center of orthogonality to the optimized tensor\cite{schollwock2011density}. %For a detailed review of MPS optimization we refer the reader 
The standard approach for training TNBMs involves a tensor-by-tensor optimization scheme, where individual tensor cores are updated sequentially. For one-dimensional tensor networks, such as matrix product states (MPS), this process is often interleaved with gauge-fixing to maintain numerical efficiency \cite{schollwock2011density}. While effective for simpler cost functions, applying this method to the negative log-likelihood (NLL) loss, central to TNBMs, reveals key limitations.

% One of the main challenges lies in the slow convergence of gradient descent (GD) for single-tensor optimization. The NLL loss, due to its logarithmic structure, requires iterative inner-loop optimizations for each tensor core, significantly prolonging the training process. This iterative process makes optimization inefficient, particularly for large-scale problems.

The tensor-by-tensor scheme has an advantage for optimization of quadratic cost function, such as expectation values of observables, where single-tensor updates can be solved directly via the eigenvalue problem. In these cases, each optimization step introduce a significant reduction of the cost function. However, for generative tasks using the negative log-likelihood (NLL) loss, the logarithmic nature of the landscape complicates optimization, requiring incremental steps of gradient descent. This introduces two key challenges: first, the inner gradient descent loop slows down training by adding a computational subroutine; second, unlike the analytical solutions in DMRG, numerical optimization is prone to nonoptimal convergence to local minima due to the abundance of singularities and minima in the NLL landscape.

In this paper, we describe an accelerated optimization scheme for TNBMs that addresses the challenges of single tensor core optimization. Our method combines an on-manifold second-order optimization approach with a regularization mechanism that prevents convergence to local minima. By leveraging second-order derivatives, the inner optimization loop converges more rapidly using fewer steps, while the regularization mechanism helps the optimizer avoid narrow minima. Our approach achieves faster convergence and lower loss values compared to gradient descent and unaltered Newton's methods, as demonstrated on both discrete (Bars and Stripes, MNIST \cite{lecun2010mnist}) and continuous-variable datasets (IRIS \cite{misc_iris_53}).

% The paper is structured as follows: Section II describes the mathematical framework of TNBMs and the challenges in their optimization. Section III introduces our regularized second-order optimization technique. Section IV presents numerical experiments and results, while Section V discusses the implications of our findings and potential extensions to more complex tensor-network architectures.

% We demonstrate our method for training a 1-dimensional MPS on two discrete datasets- the bars and stripes\cite{mackay2003information} and the MNIST dataset \cite{lecun2010mnist}, as well as training of embedded TN on the continuous-variable distribution dataset IRIS \cite{misc_iris_53}. Lastly, we verify that our method improves the generalization capabilities of the model.

This paper is structured as follows: In Section II, we introduce the mathematical framework of tensor-network Born machines and outline the challenges of optimizing their negative log-likelihood loss. Section III details our proposed second-order optimization scheme and the regularization techniques used to improve convergence. In Section IV, we present numerical results demonstrating the effectiveness of our approach on discrete and continuous datasets. Finally, Section V discusses the implications of our findings and potential directions for future work.

\section{Background}

\subsection{Tensor-network Born machines}
% Tensor-network Born machines describe distributions over bit strings using the amplitudes of a variational state, represented as a tensor network. Inspired by quantum physics, the different amplitudes are considered as the amplitude of a quantum state and can exhibit negative or complex values, while the probability of each bit string is determined by the absolute value squared of its amplitude. %in which a the state of a system is described using a list of complex amplitude. encodes a probability distribution as the amplitudes of a quantum state. 
% The quantum state is in turn represented using a tensor network, which encodes the amplitudes of a quantum state using a net of tensor cores.
% The tensor network is presented graphically by a Penrose diagram\cite{penrose1971applications} each node or box represents a single tensor core, and each edge represent an index. Indices connecting two tensors are summed over, contracting connected tensor cores into a merged tensor, and free indices function as input vertices.

Tensor-network Born machines describe distributions over bit-strings using the amplitudes of a variational state, represented as a tensor network. Inspired by quantum physics, these amplitudes correspond to those of a quantum state and can take negative or complex values, while the probability of each bit string is given by the squared absolute value of its amplitude.

Denoting the quantum state with the bra-ket notation as $\ket{\psi}$, the predicted probability of the model is given by the Born rule $p(x) = \abs{\bra{\psi}\ket{x}}^2$, where $\ket{\psi}$ is the inner quantum state of the Born machine and $\ket{x}$ is the encoding of a sample bitstring as a quantum state.

To efficiently describe the state of the Born machine, the quantum state $\ket{\psi}$ is represented as a tensor network, a network of interconnected tensor cores that reproduce global state when contracted together. This structure is commonly visualized using a Penrose diagram \cite{penrose1971applications}, in which nodes correspond to tensor cores and edges indicate index contractions. Indices connecting two tensors are summed over, merging them into a single tensor, while free indices serve as input vertices (See Fig. \ref{fig:TN_illustration}(a)).

To grade the performance of the TN, TNBM models use the negative log-likelihood (NLL) loss function. The NLL function measures how well the model's probability distribution $p(x)$ matches the observed data $\{x_i\}$, defined as
    \begin{align}
        {\rm L}(\{x_i\}) = -\sum_{x_i}{\log{p(x_i)}}.
        \label{eq:NLL-def}
    \end{align}
where $p(x)$ is the likelihood of a sample predicted by the model.
Up to a constant, the NLL function approximates the Kullback–Leibler divergence between the model and target distributions. 

In this work we focus on the one-dimensional tensor network variant known as matrix product states (MPS), which are commonly used to model the state of 1D quantum systems \cite{schuch2011classifying, cirac2021matrix}. An MPS is a one-dimensional array of tensors, each having two bond indices connecting it to the left and right, and a single free ``site'' index representing the physical state. At the edges of the system, we choose an open boundary condition, such that the first and last tensors have only a single bond index.
The dimensions of the bond indices control the level of correlation of the probability distribution between different sites, quantified by the entanglement entropy of the quantum state.% [][][].
 
For the optimization of the MPS, we look to update the different tensor cores of the state to minimize the loss function. There are several approaches to this task. A straightforward approach is to perform gradient descent on the different tensors simultaneously with a small learning rate. With this approach, in each iteration every tensor is updated slightly, often requiring hundreds or thousands of epochs to converge, which is often used for machine learning tasks. 
An alternative approach is to use local optimization scheme, where only a single tensor is updated at a time, especially when the cost function is quadratic, e.g. expectation value of a Hamiltonian. Taking advantage of tensor contraction and simplification techniques, each single-tensor optimization problem is reformulated in terms of the effective environment tensor defined as the contraction of the rest of the cores in the TN. By sweeping across all the tensor cores sequentially, the compact representation of the environment tensor can be retained and modified to describe the local optimization problem at each site (see Fig.~\ref{fig:TN_illustration}(b)), avoiding the need to repeatedly contract the entire tensor network. 
For MPS, this method forms the basis of the DMRG algorithm, widely used in condensed matter physics and quantum chemistry \cite{schollwock2011density,garnet2011density,catarina2023density}. %For further details on MPS optimization methods, we refer the reader to comprehensive reviews on the subject \cite{schollwock2011density, garnet2011density, catarina2023density}.%, and by ...[][].

For some choices of cost functions, the local optimization problems can be solved in a single step by analytically finding the minimal solution. This is the case for quadratic cost functions naturally used to describe the energy of a state in quantum systems, whose local optimization problem can be solved by singular value decomposition of the environment tensor \cite{schuch2011classifying}. For the case of the NLL cost function this is not the case, and the single tensor optimization tasks need to be solved iteratively using gradient-based optimization subroutine, slowing the optimization process.
This computational bottleneck motivates the use of optimization techniques with faster convergence for the single-tensor sub-optimization problem. To address this, we study the option of performing Newton-method optimization that operates on the constrained manifold of normalized tensors, improving convergence speed while maintaining numerical stability.

\subsection{Single tensor optimization}

Focusing on the optimization of a single tensor core, the predicted probability of a bit-string as a function of a single tensor $T$ is given by the squared overlap
\begin{align}
    p_x(T) = \abs{(T,w_x)}^2,
\end{align}  
where $w_x$ is the reduced environment corresponding to the sample $x$ (see Fig. \ref{fig:TN_illustration}(b-c)). We treat the full contraction of the two tensors as an inner product, denoted using brackets \((\cdot,\cdot)\). Assuming the MPS is in canonical form, with the center of orthogonality at \(T\), we preserve global normalization by enforcing \(T\) to be normalized with
\begin{align}
    \norm{\psi}^2 = \braket{\psi} = (T, T).
\end{align},
where the last equal sign holds because the left and right environment tensors are isometries. 

For the NLL function, the loss function for a single tensor core takes the form of  
\begin{align}
    {L}(T) = -\sum_{x}{n_x \log\left(\frac{{\left|(T,w_x)\right|}^2}{(T,T)}\right)},
    \label{eq:loss_NLL}
\end{align}  
where $n_x$ is the frequency of the string $x$ within the training set, with $\sum_{x}{n_x} = 1$. While the loss is defined for any choice of tensors, the physical interpretation of the MPS as a model of probability distribution constrains the space of valid tensors to the manifold of tensors with the constant norm of one- $(T, T) = 1$.

Geometrically, this constraint corresponds to the manifold of the hypersphere $S^{D-1}$ of dimension $D - 1$, where $D = {\rm dim}(T) = d\,\chi^2$, $d$ is the site index dimension and $\chi$ is the bond dimension of the MPS. This puts our optimization problem within the framework of optimization on Riemannian manifolds.

Following the notation used in \cite{absil2008optimization}, through this paper we use the notation \(\overline{L}(T)\) for the unconstrained loss function, and $L(T)$ for the version constrained by the constant normalization condition. %The denominator in Eq.~\eqref{eq:loss_NLL} explicitly enforces normalization, ensuring that optimization remains within the manifold of normalized states, as discussed in Section \ref{sec:optimization_algo}.  

To efficiently optimize the cost function, the constraint must be accounted for during optimization, adjusting the derivatives to be aligned with the hypersphere manifold. In the next section, we introduce a constrained Newton optimization method for TNBMs, which operates directly on the manifold of normalized tensors while enforcing this constraint.  

% \subsection{(New order- regularization)}

% We now proceed to present the three main ingredients of the optimization algorithm - second order Newton method optimizer, constrained on-manifold optimization and regularization of the landscape. These components work together to create a fast-converging algorithm tailor-made for TN- Born machine optimization.

% {\magenta New order should be:
% \begin{itemize}
%     \item Regularization first
%     \item Newton method - central results, with explanation of the on-manifold projections
%     \item modification according trust region method principles
% \end{itemize}}

% The need for regularizing the loss landscape rises from the structure of the NLL function as a sum of logarithmic terms, which exhibit a singularity when the overlap with a single sample zeros out. This singularity creates a barrier that disconnects the regions of positive and negative overlaps, which some optimizers have difficulty to trespass. 

\section{Methods}
\subsection{Second-order optimization algorithm\label{sec:optimization_algo}}

% Using second derivative can potentially improve the convergence rate of the single tensor optimization subroutine, while introducing a relatively small computational overhead.

To optimize the NLL cost function using constrained Newton's method, we first define gradients and Hessians in the context of constrained optimization. Unlike optimization of functions in free space, the derivatives of constraint functions include components that extend outside the constraint manifold. A simple gradient-based step could violate the constraint, requiring projection onto the tangent space. The gradients and Hessian tensors have to be adjusted to reside within the tangent space of the manifold, as described in the following.

% To optimize the cost function using gradient-based solvers, we calculate the gradient according to a single tensor:

The unconstrained gradient of the loss function $\overline{L}(T)$ with respect to a single tensor is given by
\begin{align}
    {\rm grad}\,{\overline{L}}(T) = 2{T} - 2\sum_{x}{n_x \frac{{w_x}}{(T,w_x)}},
    \label{eq:unconstraint_grad}
\end{align}
where the partial derivatives are taken with respect to the tensor elements.

To enforce the constraint, the gradient is projected onto the tangent space of the manifold to avoid stepping outside of the manifold, defining the constrained gradient as
\begin{align}
    {\rm grad}\,{f}(x)  = {\Pi}_{x}\left({\rm grad}\,{\overline{f}}(x)\right),
\end{align}
where $\Pi_x$ is the projection onto the tangent space at point $x$. For a constant norm constraint, the projection of a vector $v$ on the tangent space to the hypersphere at a point $x$ is given by
\begin{align}
    \Pi_x(v) = v - v_{\perp} = v - \frac{(x,v)}{(x,x)}x.
\end{align}

% Then the update rule becomes
% \begin{align}
%     T &\rightarrow T + \eta\;{\rm proj}_T(\nabla L)\\
%         %&= T + \eta\left(\nabla L - \frac{(T,\nabla L)}{(T,T)}T \right)\nonumber\\
%         &= (1-\eta (T,\nabla L))T + \eta \nabla L\nonumber
% \end{align}
% preserving the norm of the tensor up to first order in $\eta$:
% \begin{align}
%     \norm{T + \eta\;{\rm proj}_T(\nabla L)} = \norm{T} + o(\eta^2)
% \end{align}

For the loss function defined in Eq.~\ref{eq:loss_NLL}, the gradient naturally lies within the tangent space, so the projection step has no effect, i.e., ${\rm grad}\,{L}(T)  = {\rm grad}\,{\overline{L}}(T)$. For other versions of loss functions, however, this projection is important, as in the case of the regularized cost function introduced in \ref{sec:reg}.

In this work, we apply Newton’s method to accelerate the optimization of each single tensor. Newton’s method refines the optimization step by incorporating second-order derivative information, adjusting both the step size and direction based on a local quadratic approximation of the loss function. This method has been applied to tensor network optimization in the cases of fidelity and expectation value optimization \cite{lubasch2014algorithms, lubasch2018multigrid, liao2019differentiable}.
Newton’s optimizer assumes that, near a minimum, the loss function can be well approximated by a quadratic model. Using the second derivative of $L(T)$, the Newton step is give by
\begin{align}
    %T \rightarrow T - \hat{H}^{-1} \nabla{L}
     T \rightarrow T - \left({\rm Hess}\,{L}\right)^{-1} {\rm grad}\,{L}
\end{align}
where ${\rm Hess}\,{L}$ is the Hessian matrix and ${\rm grad}\,{L}$ is the gradient vector of the loss function. In free space, the Hessian is defined as the matrix of second derivatives of the loss function with respect to pairs of variables
\begin{align}
    {{\rm Hess}\,\overline{L}(T)}_{i,j} = \frac{\partial^2 L}{\partial T_i \partial T_j},
\end{align}
and for the NLL cost function, the Hessian in free space is given by the $D \times D$ matrix
\begin{align}
    {\rm Hess}\,\overline{L}(T)  = 2I + 2\sum_{x}{n_x \frac{{w_x}\otimes{w_x}}{(T,w_x)^2}}.
\end{align}
where $I$ is the identity matrix, and $\otimes$ denotes the outer product between two vectors.
% which is a $D \times D$ matrix.

For constrained optimization, Newton's method has to be adapted to avoid large jumps out of the manifold. Unlike simple gradient descent, where the constraint could be accounted for by stepping in the direction of the gradient and projecting the new point back to the manifold, the step can be large and can throw the optimizer far away from the original manifold. %Moreover, it fails to take into account the curvature of the manifold itself, which is vital for second-order approximation of the loss function on the manifold.

To properly take the constraint into account, the Hessian tensor has to be projected back to the tangent space as a tensor. Following \cite{absil2008optimization, boumal2023introduction}, 
%we describe the projected hessian tensor as a bi-linear function, which quantify the change in ${\rm D}(\nabla f)[u]$, the directional derivative in the direction $u$, along a second direction $v$ \cite{absil2008optimization}
we describe the constrained Hessian tensor as a linear map acting on the tangent space that describes the directional derivative of the gradient along a given direction %Denoted as  ${\rm D}(\nabla_T f)[u]$, this map describes the change in the directional derivative ${\rm D}(\nabla_T f)[u]$, along a second direction v
    \begin{align}
        % \hat{ H}(f)_T[u,v] = \text{proj}_T ({\rm D}(\nabla f)[u])(v) \label{eq:hess-general}
        \text{Hess}f(T)[u] = \nabla_{u} \text{grad}\,f(T).
    \end{align}
This derivative of vectoric function (or vector field) which reside on the tangent space of the manifold is named also the Riemannian connection. The same Hessian can be calculated by taking a free-space directional derivative of the constraint gradient and orthogonally projecting it onto the tangent space \cite{absil2008optimization, boumal2023introduction}
\begin{align}
        \text{Hess}f(T)[u] = \Pi_T ({\rm D}({\rm grad}\,{f})[u]). \label{eq:hess-general}
        % \text{Hess}f(T)[u] = \nabla_{u} \text{grad}\,f(T)
\end{align}
The resulting Hessian is an object which receives and returns only vectors that lie within the tangent plane, and can be written as a 2-dimensional tensor. For the particular case of NLL loss function, the projected Hessian can be written as a matrix with
\begin{align}
    {\rm Hess}\,L(T) = 2(I - T^{\otimes 2}) + 2\sum_{x}%{p_x {\frac{\ket{\text{proj}_T (w_x)}\bra{\text{proj}_T (w_x)}}{(T,w_x)^2}}}.
    {n_x {\frac{\left[\Pi_T (w_x)\right]^{\otimes 2}}{(T,w_x)^2}}}.\label{eq:NLL-Hess}
\end{align}
The expression above is obtained by taking the derivatives and projections in Eq. \ref{eq:hess-general}, and performing symmetrization on the resulting tensor,  which simplifies the expression but has no effect on the action of the Hessian on vectors inside the tangent space.

Using this projected Hessian, we can now calculate the Newton step $\Delta_N$ by solving the following set of linear equations:
\begin{align}
    \begin{cases}
        ({\rm Hess}\,L(T)) \Delta_
        {\rm N}&= -{\rm grad}\,{L}(T) \\
        (T, \Delta_{\rm N}) &= 0
    \end{cases}. \label{eq:Newton_step_eq}
\end{align}
The first equation corrects the size and direction of the bare gradient while the second equation keeps the step within the tangent plane. Notice that the Hesian is singular by definition and is no longer invertible in $\mathbb{R}^{D\times D}$.

This system of equations can be solved either by direct calculation and inversion of the full Hessian, or using iterative methods which use only individual projections of the Hessian, see appendix \ref{app:solve_newton_step} for details.

\subsection{The local-minima problem}

The constraint Newton's optimization method has potential to accelerate the convergence of the algorithm as it contains a high-order%more accurate 
description of the loss function landscape. Nevertheless, using bare Newton steps as the sole mechanism of the optimization algorithm can lead to various problems, which are manifested strongly in the case of TNBMs optimization.

\begin{figure}[t]
    % (a) NLL  \;\;\;\;\;\;\;\;\;  (b) Regularized NLL

    \centering
    \includegraphics[width=\linewidth]{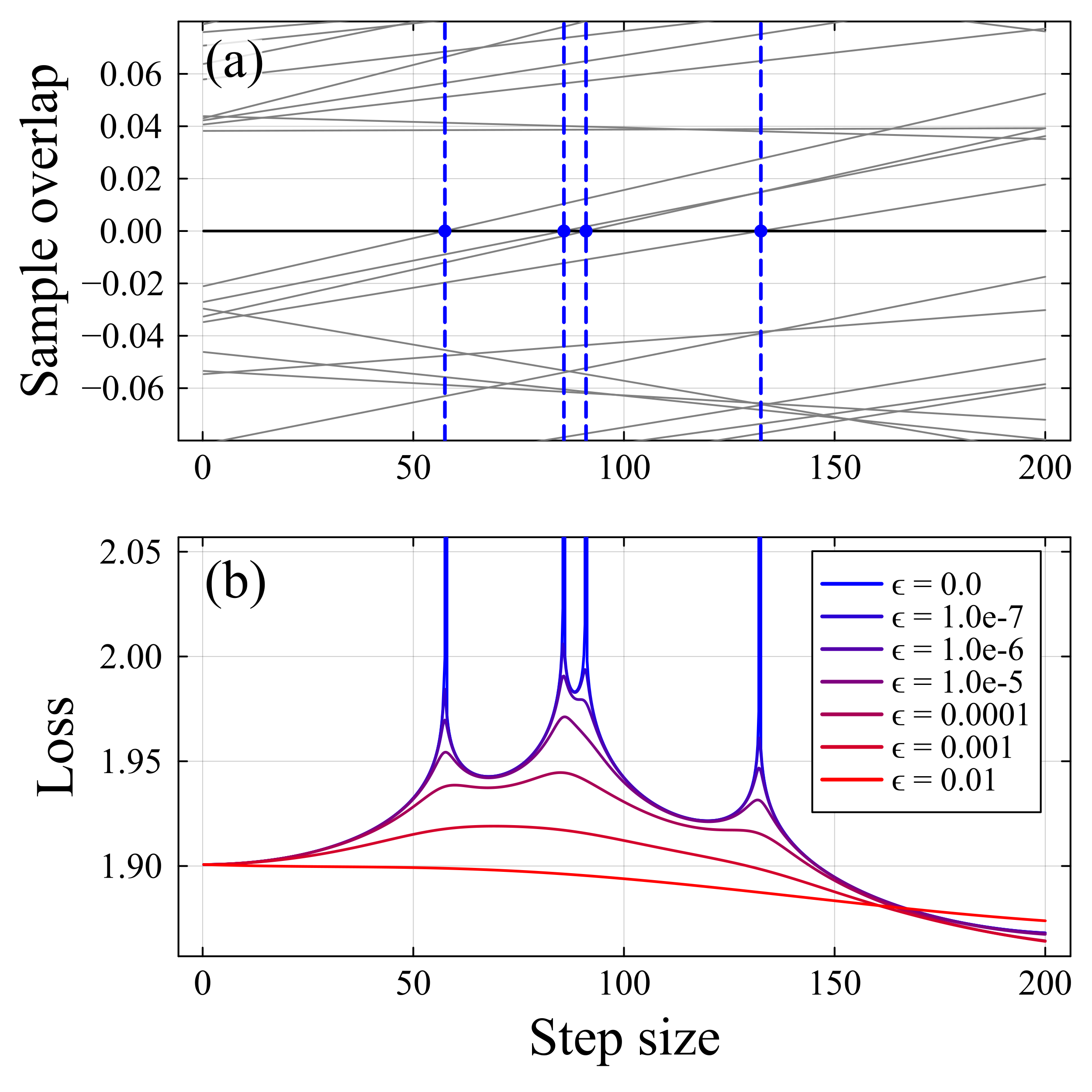}
    \caption{An illustration of the effect of regularization on the NLL loss landscapes and its relation to sample overlaps. In panel (a), the overlaps between the MPS and different samples are plotted as a function of step size along a specified direction. For small step sizes, the overlap shifts approximately linearly. In panel (b), the NLL loss landscapes are shown for varying regularization constants, plotted against the same step sizes. The loss curves are shifted vertically to align their values at the zero point for clarity. Without regularization ($\epsilon = 0$), the loss landscape exhibits barriers due to vanishing overlaps, creating challenges for optimization. The introduction of regularization smooths the landscape by introducing a cutoff, dissolving these barriers and enabling the optimizer to converge to the global minimum.  }
    \label{fig:Landscape}
\end{figure}

To understand these challenges, we examine a one-dimensional slice of the cost function, plotted in Fig.~\ref{fig:Landscape}. 
As illustrated in Fig.~\ref{fig:Landscape}(b), the landscape is filled with many asymptotic singularities, creating many narrow local minima. 
This structure persists in the multidimensional tensor space: the singularities appear across linear planes that dissect the landscape into many small regions, each with its distinct local minima. This is a natural property of the NLL loss function, as it is constructed from many logarithmic poles, one for each sample (see Eq.~\ref{eq:loss_NLL}), and the overlap of the tensor $T$ with the different samples create singularity walls on the manifold that cannot be transversed using a continuous gradient flow. The logarithmic poles dissect the landscape into $2^{N_s}$ regions, corresponding to all possible configurations of overlap signs.

As a result, the choice of optimizer plays a key role in successful convergence. 
With steepest descent, the optimizer naturally avoids narrow minima by taking large steps and leaping away from narrow gorges with steep walls. This overshooting mechanism filters potential local minima with high second derivatives and converges only to points that are relatively flat. 
In contrast, Newton’s method adjusts both the direction and magnitude of each step using the Hessian, and the optimizer tends to jump directly into one of the closest minimal points, adjusting the size of the step to avoid overshooting. 

To discourage the convergence to very narrow local minima, while still benefiting from the fast convergence of Newton's method, we introduce regularization of the loss function that smooths out the landscape and cuts off the logarithmic divergences.

\subsection{Regularization of the NLL loss landscape \label{sec:reg}}

In order to soften the barriers created by the logarithmic poles we introduce a regularization mechanism for the NLL loss landscape. 
We explore two types of regularization, each offering a different approach to mitigating singularities: one that smooths the landscape by spreading out singularities locally, and another that introduces a bias toward positive overlaps. We discuss the first type in this section, and the second one in the next section, in the context of continuous-variable models. 

In the first regularization variant, we add a constant to the argument of each logarithm, broadening and smoothing the singularities of the landscape. Given the regularized log function
\begin{align}
    l_\epsilon(x) = \log\left( \abs{x}^2 + \epsilon\right),
\end{align}
where $\epsilon$ is a positive regularization constant, we define the regularized cost function as the sum of the broaden logarithms of the overlaps with
\begin{align}
    % L_\epsilon(\psi) = -\sum_x{n_x \log\left( \abs{\bra{\psi}\ket{x}}^2 + \epsilon\right) },\label{eq:reg}
    L_\epsilon(\psi) = -\sum_x{n_x l_\epsilon(\bra{\psi}\ket{x}) }.\label{eq:reg}
\end{align}

\begin{figure}[t!]
    \centering
    \includegraphics[width=1\linewidth]{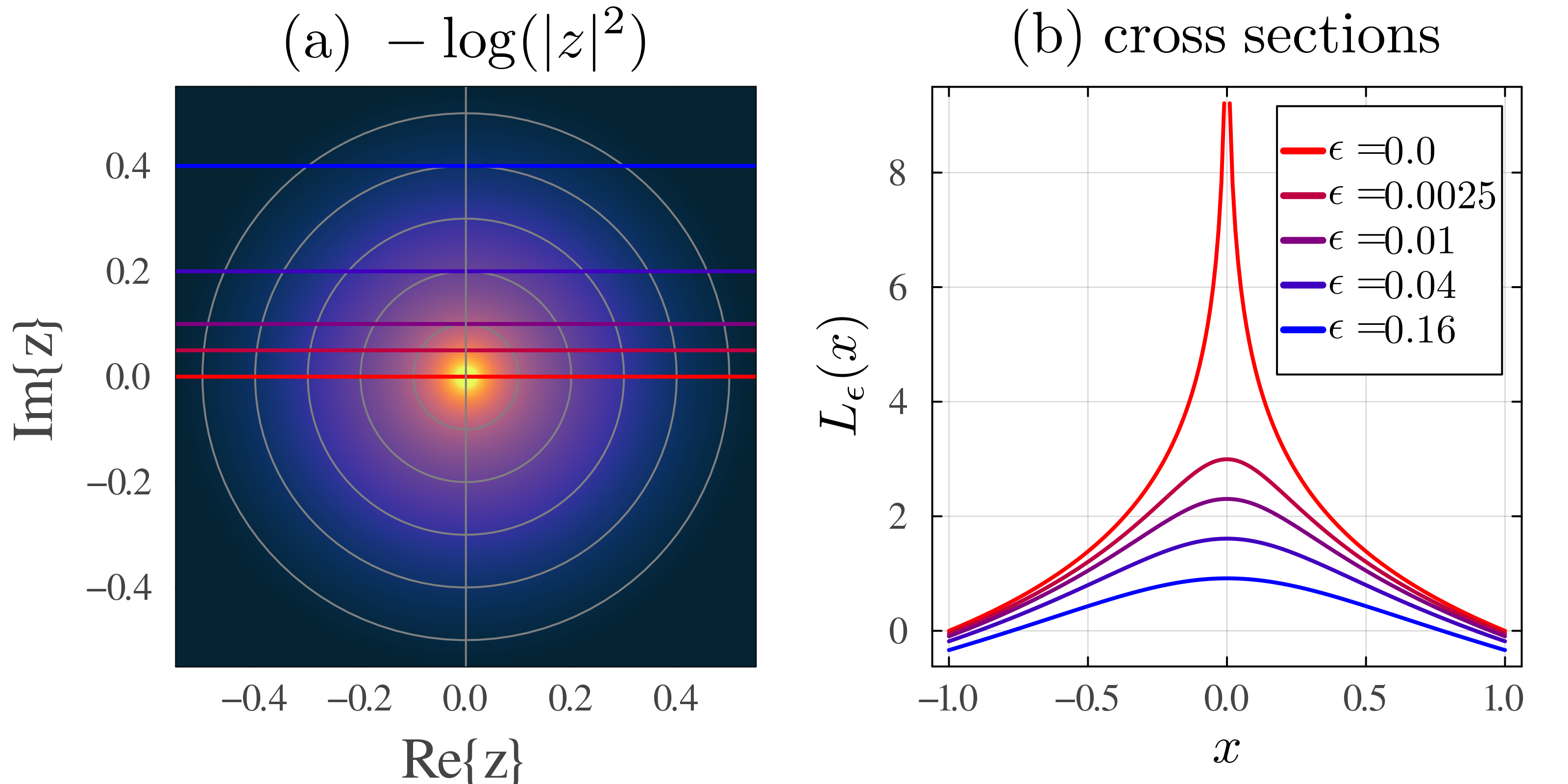}
    \caption{An illustration of the regularization as an imaginary shift in the complex plane. In panel (a), the negative logarithm function is extended to the complex plane, accounting for overlaps with complex phases. By introducing a constant imaginary shift to the overlap, the singularity can be avoided. In panel (b), we plot 5 cross sections for different imaginary shifts, effectively smoothing-out the singularity at the zero overlap point.}
    \label{fig:reg_complex}
\end{figure}

The regularization constant keeps the argument of the $\rm log$ function positive even when the overlap between the state vector and the sample is zero. Effectively, this type of regularization introduces a constant imaginary part to all overlaps, keeping their norm above zero even when the bare overlap vanishes. This is a cost-effective way to circumvent the divergent pole in the complex plain with real-valued tensors, as illustrated in Fig.~\ref{fig:reg_complex}.

An alternative interpretation of the effect of regularization is as a smoothing, or filtering, of the original loss landscape using a convolution with a kernel function. Choosing a kernel function of a Lorentzian 
\begin{align}
    \kappa_\epsilon(x) &= \frac{1}{\pi}\frac{\sqrt{\epsilon}}{x^2+\epsilon},
\end{align}
with a width of $\gamma = \sqrt{\epsilon}$, the convolution of the kernel function with the logarithmic pole recreate the regularized NLL from Eq. \ref{eq:reg}
\begin{align}
    % L_\epsilon(\psi) &= \left(\mathcal{L}_0(x) * \kappa_\epsilon(x)\right)|_{\bra{\psi}\ket{x}}.
    l_\epsilon(x) &= l_0(x) * \kappa_\epsilon(x).
\end{align}

The smoothing of the NLL function prevents sharp gradients and singularities on one hand, while producing an easy-to-calculate closed formula at the same time. Another important effect of regularization is a reduction in the number of local minima across the landscape. %When each pole is smoothed out simultaneously using a large regularization constant, sharp features are washed out, and by increasing the regularization constant more and more local minima are eliminated, as plotted in Fig. \ref{fig:Landscape}. On the same time, the position of the minimal points tend to stay in place with only minimal shifts in the corresponding $x$ value. 
As shown in Fig. \ref{fig:Landscape}(b), increasing the regularization constant smooths out sharp features of the landscape, effectively eliminating local minimum points as the constant grows. Meanwhile, the positions of the remaining minima remain largely stable, with no significant shifts in their positions.

Using the regularized loss function, we can reapply Newton's optimization method and avoid local minima while benefiting from the accelerated convergence of the second order optimization.

The implementation of Newton's method on the new regularized loss function requires caution, as several properties of the cost function have changed: the projections of the gradient and Hessian have to be readjusted according to the new regularized loss function (see Appendix \ref{app:reg_derivs}). In addition, the smoothing of the cost function replaces many singularities with local maxima points with concave regions around them (see Fig. \ref{fig:Landscape}(b)). 
This has a negative impact on the convergence of Newton's optimizer, which rely on the convexity of the function, and can otherwise converge to any critical point, failing to distinguish between maxima and minima.

% In order to optimize the cost-function in regions where the cost function is non-convex, we take an approach inspired by the trust region method, which bounds the norm of a single step to reside in a region where the polynomial approximation is valid. In case there is no minimal point within the interior of defined trust region, the optimizer searches for a local minimum on the boundary of the region, shaped either as a constant-radius hypersphere or, more generally, as a hyperboloid.

To address this, we take an approach inspired by the trust region method \cite{more1983computing, conn2000trust, yuan2000review, adachi2017solving}, which mitigates the issue by bounding the size of the optimization step. %In the trust-region method, the optimization step is determined using a quadratic approximation (or model) of the function, minimizing this model within a designated \textit{trust region}— a region where the approximation is assumed to be valid. 
In the trust-region method, the optimizer takes into account that the second order approximation is expected to be valid only within a bounded region of the search space (the \textit{trust region}).
As a result, the optimization step is bounded to either the interior of the region, if a local minimum exists within it, or to the boundary otherwise.

In cases the model is convex and a minimum is available inside the trust region, the step coincides with the Newton step. When there is no minimal point within the interior of the defined trust region, the optimizer searches for a local minimum on the boundary of the region, shaped either as a constant-radius hyper-sphere or, more generally, as a hyperboloid. %By enforcing this constraint, the method enforces from second-order corrections even in nonconvex regions while maintaining step-size control, ensuring stable optimization.
By enforcing this constraint, the method ensures stable improvement at each step while benefiting from second-order corrections even in nonconvex regions.

The minimum on the boundary of the region is found by adding Lagrange multipliers as  additional quadratic terms to the local minimization problem, setting a new constraint over the size of the step taken. 
The additional Lagrange multiplier introduces a shift in the Hessian of the model, with either a scalar for an isotropic trust region or, generically, with a positive Hermitian matrix, whose magnitude is determined by the constraint over the step size. This effective shift eliminates all the negative eigenvalues of the Hessian, enabling the optimizer to find the point with the lowest loss value on the boundary, rather than jumping into a false critical point. 

In the case of the regularized NLL landscape, we can apply the principles of the trust region algorithm and correct the negative parts of the Hessian by adding quadratic terms as additional constraints. In our case, rather than adding a full scalar matrix and inhibiting the rate of convergence in all directions equally, we add terms that directly counter only the negative subspace of the Hessian. 
This can be done by taking the absolute values of all the terms that contribute to the negative part of the Hessian, which ensures that the modified matrix stays positive with minimal modification (see Appendix \ref{app:reg_derivs}), which is equivalent to constraining the Newton step only in directions where the point is approaching a logarithmic singularity. 
Unlike the trust region method, we do not determine the size of the trust region
apriori, rather we accept the step size resulted from this heuristic, although it is possible to expand our algorithm to include a precise control over the step size. The approach of adding additional terms as a numerical damping of the Hessian is also common in other numerical optimization algorithms, such as the Levenberg-Marquardt algorithm \cite{more2006levenberg} and the Tikhonov regularization procedure in regression problems \cite{golub1999tikhonov}.
% To fix this problem, we artificially modify the Hessian by taking the absolute value of all terms in Hessian formula, which ensures that the modified matrix stays positive.

\subsection{Second variant- regularization by shift}

A second version of regularization mechanism is relevant in cases when we can infer the preferable configuration of amplitude signs a priori. In the case of site indices that represent a digitization of a continuous variable, as a particular example, we expect all related amplitudes to share the same sign, due to the continuous nature of the original variable. Fitting the same function with alternating signs can cause overfitting of the data, and impact the overall performance and generalization capabilities of the model.
In such a case, we can introduce a bias-regularization to favor the same signs for all amplitudes by the addition of a constant to the overlap amplitude itself
\begin{align}
    L^{\rm bias}_{\epsilon_b}(\psi) = -\sum_x{p_x \log\left( \abs{\bra{\psi}\ket{x} + \epsilon_b}^2\right) } \label{eq:reg_bias}
\end{align}
The effect of this bias on the optimizer can be viewed as moving all singularities to the negative side of the amplitude, therefore favoring positive amplitude rather than negative ones. 
Unlike the first variant of regularization, increasing the bias does not smooth the sharp singularities of the landscape, but rather shifts it in such a way that increases the region connected to positive overlaps, encouraging the optimizer to converge to the solution of uniform-sign overlaps.

\section{Results}
% \begin{itemize}
%     \item technical details about cutoff schedules 
%     \item Numerical results for training on known datasets- BAS, MNIST ect. 
%     \item comparison with simple steepest descent, complex valued gradient descent, and some other alternatives
%     \item Comparison of the necessity of regularization for various bond-dimension values. \ref{BAS}
%     \item An example with continuous-embedding MPS? \ref{fig:continuous}
%     \item Examples for improved generalization \ref{fig:generalization}
% \end{itemize}

To test our new optimization algorithm, we perform a comparison between simple gradient descent, Newton's method, and the regularized Newton optimization algorithms on several datasets.  

% To demonstrate the utility of the regularized Newton's method optimization algorithm 
We train an MPS tensor network on two datasets- the bars and stripes dataset (BAS) \cite{mackay2003information} and the MNIST dataset \cite{lecun2010mnist}. The bars and stripes dataset is a synthetic dataset consisting of binary $n \times n$ grids, where each cell in the grid is either 0 (black) or 1 (white). The pictures in the dataset consists of either ``bars'' (columns), or ``stripes'' (rows) of black and white pixels. The MNIST dataset is a dataset of hand-written digits, widely used in machine learning and computer vision, mainly for benchmarking classification algorithms. The dataset contains gray-scale images of 28 by 28 pixels of handwritten digits.

We use these two dataset to compare the different optimization algorithms discussed in this work- steepest descent, Newton's method, and the regularized Newton's method by smoothing of the landscape. For both the BAS and the MNIST datasets, we prepare datasets of images with 7 by 7 grid. %For the MNIST dataset, we course-grain the original grid of 28 by 28 pixels to 7 by 7 pixels and set a threshold to convert it from grayscales to a binary image.
For the MNIST dataset, we perform downsampling on the original 28 by 28 grid via average pooling, then binarizing the result using a threshold.
We represent the final 2-dimensional grid of 49 binary pixels using a 1-dimensional MPS, unraveling the pixels for the MNIST dataset in a snake-like configuration to retain as much of the original notion of locality as possible in the transition from 2 to 1 dimensions.  We then scramble all subgroups of different labels together, drawing random samples from arbitrary labels as the training set.  

Implementing our new optimization algorithm, we define a regularization schedule that varies the regularization constant throughout the optimization. By starting with relatively large values of $\epsilon$ that decrease with each iteration, we smooth out peaks and explore the whole landscape without barriers and avoid getting stuck in local minima early at initial iterations, while maintaining good optimization accuracy later when the optimizer reaches the proximity of the desired global minimum.
In our numerical experiments, we chose an exponentially decreasing regularization schedule, starting from a moderate value of $\epsilon = 0.025$ and scaling down exponentially for later sweeps.

\begin{figure}[t]
    \centering
    \includegraphics[width=1\linewidth]{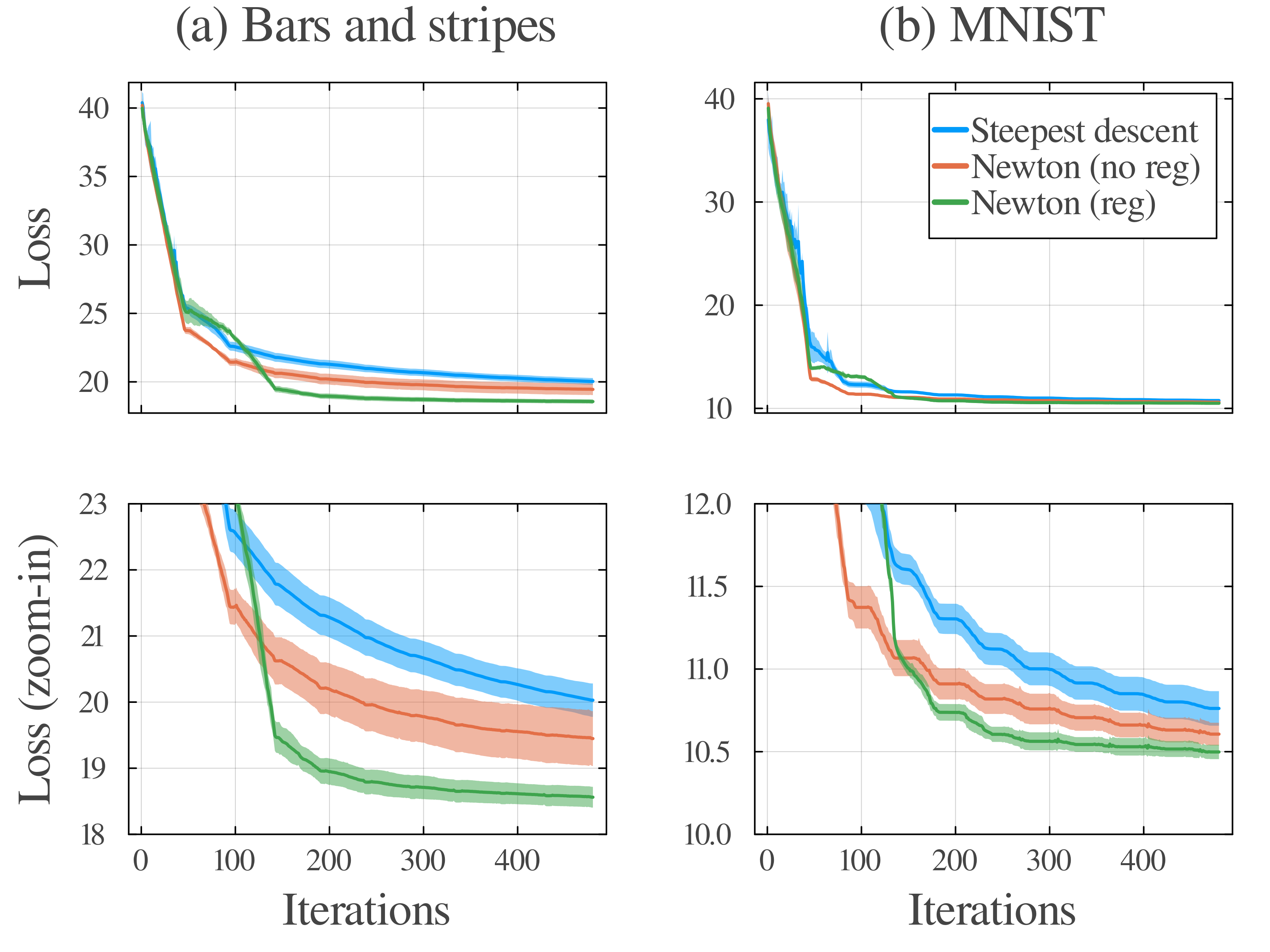}
    \caption{A comparison of loss curves for training of TNBM on the bars and stripes (BAS) and the MNIST datasets using different optimization algorithms. We compare the regularized Newton method (green) %our method 
    against the steepest descent (blue) and the vanilla Newton's method (red). The figure presents the loss as a function of the number of iterations, while the bands represent the standard deviation of the loss curves between five randomly initialized realizations. The bottom row presents a zoom-in view on the region of lower loss values reached after a single forward and backward sweep.}% (PLACEHOLDER) \jc{For this image, we're highlighing that in small $\chi$ regions, the regularization would be very beneficial. The large $\chi$ region may not always be our target, due to the overfitting or other considerations. For this figure, we can highlight one of these four. }}
    \label{fig:BAS_MNIST}
\end{figure}

For both datasets, we set the tensor network as a 49-sites MPS with a bond dimension of $\chi = 5$, and optimize it using 5 forward and backward sweeps. We train our model using 100 samples from the dataset, and average the results over 5 random initialization of the model. 
For each of these optimization algorithms, we iterate across all tensors back and forth, taking a single optimization step per tensor. 
Figure \ref{fig:BAS_MNIST} presents the resulting loss curves of the TNBM model as a function of the number of iterations, for both the BAS and the MNIST datasets. In the figure we compare the loss curve of our regularized optimization algorithm with simple gradient descent and with an unaltered version of Newton's method. The loss score metric is presented in its original form for all methods, i.e. them non-regularized loss as written in Eqs.~\ref{eq:NLL-def} and \ref{eq:loss_NLL}. % WRITE: [100 samples, 49/49 sites, GD rate- 0.1, 5 repetitions, sched_factor = 5 -> epsilon = 0.025 with relaxation of 10.^range(-1,10,12) between half-sweeps]
We observe that the regularization mechanism improves the quality of the converged model, reaching lower loss values faster compared to gradient descent or the naive newton optimizer. 

The loss curve reveals different stages of optimization of the regularized Newton optimization method. In the first sweep, for about 50 iteration, the loss curve of the regularized optimization follows the same trend of both Newton and gradient descent optimization. By the second sweep, the differences between optimizers start to emerge, with the unaltered constrained Newton's method shows an improvement over both gradient descent and the regularized Newton. At this stage the regularization increases the NLL score of the model, as the optimizer ignores the effect of nearby singularities. As the regularization constant diminishes, the bump in the loss curve of the regularized optimization flattens, and the regularized optimizer reaches lower loss values at early iterations, compared to the other optimizers. 
There are several differences between the results of the two datasets. For the BAS dataset, the optimization curve is smooth, with diminishing returns for subsequent sweeps. For the MNIST dataset, the optimization curve after the first sweep is structured as a descending staircase- a sequence of downward slopes interleaved with plateau sections. This jagged structure is due to the difference in pixel variance between peripheral and central pixels along the snake-like path n the two dimensional grid. 

The comparison between the optimizers for the MNIST dataset shows a consistent, though less significant, advantage compared to the results with the BAS dataset. The reason behind this difference is likely related to the complexity of the MNIST dataset, containing 10 different labels digit labels, as well as a more complex two dimensional geometric structure that is hard to capture using a one dimensional MPS. Those properties increase the difficulty of optimization for all models equally, which lead to closer results between the three of them.

Next we test our method for training a model of continuous-variable Born machine model \cite{meiburg2025generative}, designed to model a probability distribution over continuous variables using tensor networks. The model is structured similarly to the TNBM model, with an additional layer of tensors applied to the site indices of the MPS that digitizes and embeds the continuous variable input into a discrete low dimensional feature space. This additional layer includes two components- an embedding mechanism that uses a fixed mapping to transform a continuous input data into a high-dimensional feature space, and a dynamic isometric tensor which is used to extract the essential features of the data and reduce the dimensionality of the data encoded by the MPS. 
The optimization algorithm for the continuous-variable Born machine is performed similarly to the optimization algorithm of the discrete TNBMs, with the additional training of the dynamic embedding tensors, which is done using gradient descent (see \cite{meiburg2025generative} for additional details).

\begin{figure}[t!]
    \centering
    \includegraphics[width=\linewidth]{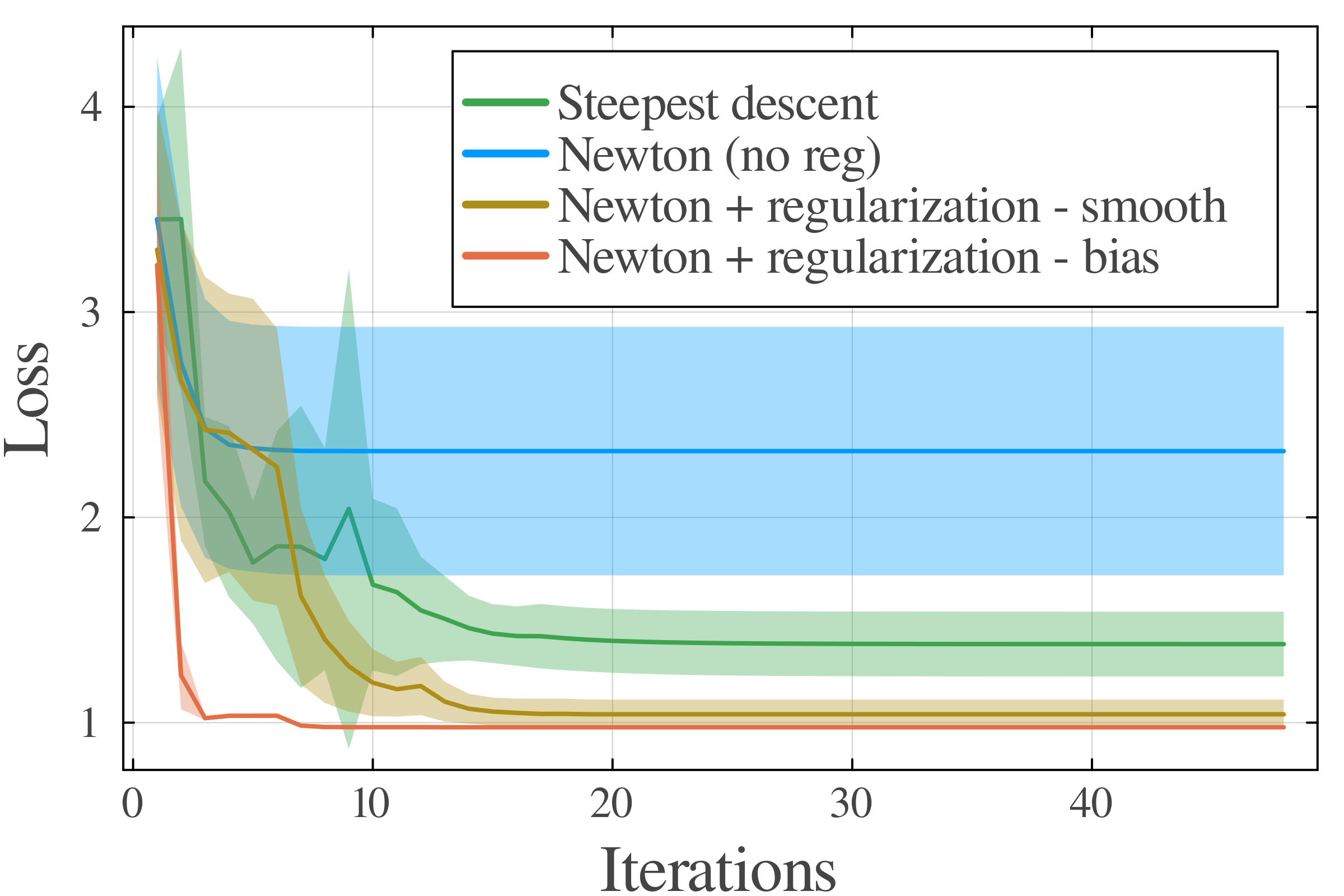}
    \caption{Loss curves of continuous embedding MPS for various optimization methods:  simple gradient-descent, Newton's method optimization and the two variants of regularized newton optimization on manifold, the smoothing regularization in Eq.~\ref{eq:reg} and bias regularization in in Eq.~\ref{eq:reg_bias}. }
    \label{fig:continuous}
\end{figure}

For the continuous model, we train a continuous Born machine on the Iris dataset \cite{misc_iris_53}. 
The Iris dataset contains 50 samples of Iris flowers, describing 4 features of each flower such as the fetal and petal length and width, as well as its species. We model the distribution using a 4-site one-dimensional TN, with a dynamic feature dimension with an initial feature dimension of 25, converted to a low feature dimension of 3. The results are plotted in Fig.~\ref{fig:continuous}, comparing the loss curve for the different optimization methods, averaged over 5 realizations. We compare the following optimization methods: simple gradient descent with $\eta = 0.05$, unaltered Newton-method optimization, and two versions of regularized Newton's method - using smoothing and shifting of the singularities.  
Comparing the unaltered Newton's method to the simple gradient descent, it is clear that in the continuous case the Newton optimizer quickly converges to a local minimum with high loss value. The variance between realizations for the Newton's method is also large, hinting to an inconsistent quality of convergence, converging to various different local minima. Adding regularization to the landscape addresses the problem of local minima convergence, and the loss of both regularization variants converge consistently to low loss values, improving upon the steepest descent optimizer. 
The results using the biased-regularization version of the optimizer show a very sharp convergence to the optimal value during the very first sweep of iterations. This is not usually the case for datasets of discrete variables such as the BAS and the MNIST datasets.  A potential reason for this behavior is that the continuous embedding naturally prefer mapping samples of continuous variables amplitudes of the same sign, making the pattern of the continuous distribution more apparent and easy to learn.

% The significant improvement of the convergence of the regularized Newton optimizer over gradient descent for the continuous MPS model has also a significant effect on the predictive capabilites of the model and for yielding better generalization for the train data. In Fig.~\ref{fig:generalization}, we present a map of the train and test error for the same dataset for varying sizes of bond and feature dimensions. With increasing bond and feature dimensions, the model gains more degrees of freedom and is prone to overfitting over the training data. We compare the train error and the test error for the optimized tensor-network using gradient descent optimization and regularized Newton's method. 
% For regularized Newton's method the test error keeps improving for large feature values and bond dimensions...

% \begin{figure}
%     \centering
%     \includegraphics[width=0.95\linewidth]{plots/generalization_comparison.png}
%     \caption{The loss value of continuous MPS optimization with and without regularization. Using regularized loss function results in better generalization and gives lower loss value for samples in the test dataset. (PLACEHOLDER)}
%     \label{fig:generalization}
% \end{figure}

\section{Conclusion and discussion\label{sec:conculsion}}

\begin{table}[t]
\centering
\begin{tabular}{|c|c|c|c|c|}
\hline
 & & \textbf{Gradient} & \multicolumn{2}{c|}{\textbf{Newton's Method}} \\ \cline{4-5}
 & & \textbf{Descent} & \textbf{Dense} & \textbf{Sparse}\\ \hline
\multirow{2}{*}{\textbf{grad}} & \textbf{time} & $N_s D$ & $N_s D$ & $N_s D$ \\ \cline{2-5}
 & \textbf{memory} & $D$ & $D$ & $D$ \\ \hline
\multirow{2}{*}{\textbf{Hess}} & \textbf{time} & -- & $N_s D$ & $N_s D$ \\ \cline{2-5}
 & \textbf{memory} & -- & $D^2$ & $D$ \\ \hline
\multirow{2}{*}{\textbf{step}} & \textbf{time} & $N_s D$ & $N_s D + D^3$ &
    \begin{tabular}{@{}c@{}}
    $N_s D N_{\text{iter}}$ \\
    $N_{\text{iter}} \sim \log(1/\varepsilon)\sqrt{\kappa}$
    %\frac{\log(1/\varepsilon)}{\log(\frac{\sqrt{\kappa}+1}{{\sqrt{\kappa}-1}})}$
    \end{tabular} \\ \cline{2-5}
 & \textbf{memory} & $D$ & $D^2$ & $D$ \\ \hline
\end{tabular}
\label{tab:complexity}
\caption{Complexity comparison between gradient descent and Newton's method optimizers, for calculation of gradients, Hessians and full iteration step.}
\end{table}

In this work we introduced an improved optimization algorithm for tensor-network Born machines, capable of reaching high quality convergence faster than simple gradient descent while exploring the global manifold and avoiding local minima. 
The algorithm utilizes restricted version of the Newton optimization method on manifold, together with a regularization of the loss function landscape to mitigate singularities. We described two versions of regularization, introducing either smoothing or shifting of the logarithmic singularities, to help the optimizer avoid local minima and converge to a lower loss point.  We have shown that this combination robustly find better quality minimal solutions faster than gradient descent, demonstrating the effect for the discrete datasets of bars-and-stripes and MNIST, and for the continuous IRIS dataset with a continuous TNBM.
Our algorithm can be naturally expanded and implemented for various tensor-network architectures, such as two-dimensional PEPS \cite{verstraete2004renormalization, orus2014practical} and the tree-like MERA network \cite{vidal2007entanglement, evenbly2009algorithms}, which can improve model performances for datasets with different geometrical properties.

The effect of regularization on the landscape mimics the effect of complex-valued tensors by effectively getting around the logarithmic pole with a constant imaginary shift of the overlap, while avoiding the memory and computational overhead of handling complex matrices. This approach restores the feasibility of using real-valued tensors for optimization, enabling efficient training without the need for explicitly introducing complex-valued representations.

A potential concern with our method is the computational cost of Newton’s optimization, particularly the need to calculate the Hessian. However, for the case of TNBM, we can use the structure of loss function to avoid explicitly computing and storing the full Hessian matrix. Instead, we use iterative solvers that only require Hessian-vector products. Instead of constructing the Hessian matrix out explicitly causing huge computational and storage waste, it is more efficient to use iterative solvers for the Newton step, which only require evaluating the projection of the Hessians on vectors, significantly improving efficiency.%lazy evaluate by listing out all contraction steps and contracting in an optimal order, significantly improving efficiency. 
For the NLL loss function, both single-tensor gradient and Hessian calculations involve only \( O(D N_{\rm s}) \) operations, for tensor dimension \( D \) and \( N_{\rm s} \) samples, with memory complexity of \( O(D) \) for gradients and \( O(D^2) \) if storing the full dense Hessian were necessary. Similarly, Newton's optimization step can be computed by either a direct matrix inversion or by using iterative methods.

% The complexity of our algorithm can be estimated as a special case of the complexity of Newton's method while taking into account the sparse representation of the derivatives of the loss function. For the NLL loss function, both single-tensor gradient and Hessian calculations involve only $O(D N_{\rm s})$ operations, for tensor dimension $D$ and $N_{\rm s}$ samples, with memory complexity of $O(D)$ for gradients and $O(D^2)$ to store the full dense Hessian.
% Comment: the cost of a single iteration also includes other heavy overheads independent on the local optimizer- such as calculation of the environment tensor and singular value decomposition, both with time complexity roughly within the range of $O(D^2)$ and $O(D^3)$.
Solving Newton's linear equation with dense matrix representation takes about $O(D^3)$ operations, while by using iterative solvers such as the conjugate gradient method \cite{hestenes1952methods, shewchuk1994introduction} or the minimal residual method (MINRES) \cite{paige1975solution, saad2003iterative}, the complexity can be reduced to $O(D N_{\rm s} N_{\rm in-iter})$ by limiting the number of inner iteration of the linear equations to $N_{\rm in-iter}$. The number of iterations required to reach accuracy of $\epsilon$ can be estimated in turn for the iterative algorithms using convergence theorems \cite{polyak1987introduction, hackbusch1994iterative, liesen2004convergence}, with the worst case estimated as $N_{\rm iter} \sim \log(1/\varepsilon) \sqrt{\kappa}$ for conjugate gradient descent, as an example.
The comparison is summarized in Table \ref{tab:complexity}.
The complexity analysis demonstrate that Newton’s optimization can be performed efficiently using iterative solvers, achieving second-order convergence with only a small computational overhead compared to first-order methods.  

% DISCUSS THE EFFECT OF BOND DIMENTION AND NUMBER OF SAMPLES

It should be noted that the regularization mechanism does not hold as a general-purpose solution for local minima trapping in TNBMs. The general task of training a generative model using limited resources is a challenging task, and naturally, the quantum inspired approach taken by the TNBM can at times have difficulties to converge to a good solution due to various reasons, such as  incompatibility of the chosen geometry and dimensions of the tensor network, and the lack of sufficient training data or computation time, as well as multi-body patterns that are hard for the model to recognize. Many times these problems manifest as convergence to local minima, often by effects distributed along many tensor cores, which cannot be solved by regularization. 
Despite these limitations, our optimization method provides significant acceleration in many cases, enhancing the efficiency of TNBMs for real-world applications. % By integrating second-order optimization with targeted regularization, we mitigate some of the key difficulties in non-convex loss landscapes, enabling better convergence in practical scenarios

% OTHER BOTTLENECKS - lower bond dimension contraints may "ruin" the large Newton step

% \begin{itemize}
%     \item outlook about the combined effect of all the components we used
%     \item Discussion on the pros and cons of using real-valued tensor cores, compared to complex tensors
%     \item thoughts about the validity of our results for high dimensions, different number of samples
%     \item complexity analysis? or discussion on the tradeoff in our algorithm (computation time per iteration, memory usage)
%     \item using out method for TN other than MPS
%     \item effects of extreme regularization
%     \item The advantage of the existance of phases in Born machines
%     \item the limits of high-order local optimizers (region of convergence and poles)
%     \item more future directions and analysis - connection between low convexity and NLL, and the approximate stability of the minimum point for multiple valu 
% \end{itemize}

\section{Acknowledgments}

This project was initiated as an internship project at Zapata AI during Summer 2023, which since then has ceased operation.  
We would like to thank Emanuele Dalla Torre from Bar-Ilan University and the Zapata AI team: Alejandro Perdomo-Ortiz, Martha Marui, Vladimir Vargas-Calderón, and Artem Strashko for the many discussions and helpful suggestions throughout our project.
This research is supported by the Israel Science Foundation, grants number 2126/24 and 2471/24.

\appendix
\begin{widetext}
\section{Derivation of the regularized gradient and Hessian \label{app:reg_derivs}}

Once we modify the loss landscape by adding regularization terms, the projections have to be adjusted as well. 

The free-space derivative of the regularized NLL is 
\begin{align}
    {\rm grad}\,\overline{L}_\epsilon(T) = 2T - 2\sum_x{p_x \frac{(T,w_x) w_x}{(T,w_x)^2 + \epsilon}}
\end{align}

%For convenience, we define the
Note that when $\epsilon$ approaches zero the gradient goes back to Eq. \ref{eq:unconstraint_grad}.

The updated gradient is no longer parallel to the tangent space. The projected gradient takes the form of:
\begin{align}
    \text{grad}\, L_\epsilon(T)  &= \text{proj}_T\left({\rm grad}\,\overline{L}_\epsilon\right) = {\rm grad}\,\overline{L}_\epsilon - \left({\rm grad}\,\overline{L}_\epsilon,T\right) T \\
    &=  2\sum_x{p_x \frac{(T,w_x)^2}{(T,w_x)^2 + \epsilon}} - 2\sum_x{p_x \frac{(T,w_x) w_x}{(T,w_x)^2 + \epsilon}}\\ 
    &= - 2\sum_x{p_x \frac{(T,w_x) \text{proj}_T(w_x)}{(T,w_x)^2 + \epsilon}}
\end{align}

Likewise, the Hessian is calculated by taking the free-space derivative of the projected gradient, and again projecting the resulted tensor back to the manifold.

The derivative of the gradient is:

\begin{align}
    \frac{\partial}{\partial T}\text{grad}\,\overline{L}_\epsilon(T) = 2\sum_x{p_x\frac{(w_x,T)\left((T,w_x)I + T\otimes w_x\right)}{{|(T,w_x)|^2 + \epsilon}}} + 2\sum_x {p_x \frac{|(T,w_x)|^2-\epsilon}{\left(|(T,w_x)|^2 + \epsilon\right)^2} {\rm proj}_T (w_x) \otimes w_x}
\end{align}
and the final projected Hessian becomes
\begin{align}
    {\rm Hess}\,\overline{L}_\epsilon(T) = 2\sum_x{p_x\frac{|(T,w_x)|^2\left(I - T^{\otimes2}\right)}{{|(T,w_x)|^2 + \epsilon}}}  + 2\sum_x {p_x \frac{|(T,w_x)|^2-\epsilon}{\left(|(T,w_x)|^2 + \epsilon\right)^2} \left({\rm proj}_T (w_x)\right)^{\otimes 2}} 
\end{align}
Again, when $\epsilon$ approaches zero the Hessian goes back to the original expression in Eq. \ref{eq:hess-general}.

As expected, for large values of $\epsilon$ some terms of the regularized Hessian can potentially become negative, creating regions where the landscape in no-longer convex. To counter this problem we introduce an aritficial correction to the Hessian and take the absoloute value of all terms, flipping the sign of the negative terms: 
\begin{align}
    \hat{H}_\epsilon^{\rm abs} = 2\sum_x{p_x\frac{\abs{(T,w_x)}^2\left(I - T^{\otimes2}\right)}{{\abs{(T,w_x)^2} + \epsilon}}}  + 2\sum_x {p_x \frac{\abs{\abs{(T,w_x)}^2-\epsilon}}{\left(\abs{(T,w_x)}^2 + \epsilon\right)^2} \left({\rm proj}_T (w_x)\right)^{\otimes 2}} 
\end{align}

This effectively eliminates the tendency of converging to local maximum points.

\section{Finding Newton step \label{app:solve_newton_step}}
In the main text, we have derived a set of equations to solve for Newton step calculation (see eq. \ref{eq:Newton_step_eq}). 
Combined together, the linear set of matrix equations can be rewritten using block matrix manipulations as
\begin{align}
            (\hat{H}^2 + 4 T T^\dagger) \Delta x = \hat{H} \nabla L
\end{align}
This set of equation is easier to solve- as it avoids the singularity of the Hessian outside of the tangent space while including implicitly the constraint on the gradient.

Solving these linear equations can be done in two ways, each with its own numerical advantages. The first method is the direct one, calculating and storing in memory the entire Hessian and then solving the set of equations using a direct linear equation solver, such as Gaussian elimination for example. With the direct method, the complexity for each iterations grows cubically with the size of the matrix as $O(D^3)$, which may become problematic for tensors with relatively large bond-dimensions.

Another approach is to solve the system of equations iteratively, each time calculating only a projection of the Hessian on single vector and saving memory in case the dimensions of $T$ are large. We have used the krylov subspaces algorithm using the krylovkit.jl package \cite{Haegeman_KrylovKit_2024}. A discussion about the complexity of the different approaches is detailed in Section \ref{sec:conculsion}.

% \section{Regularized Newton method and the trust region problem}

\end{widetext}
\bibliography{bibliography}

\end{document}